\documentclass[letterpaper, 10 pt, conference]{ieeeconf}
\IEEEoverridecommandlockouts
\usepackage{cite}
\usepackage{booktabs} %
\usepackage{cite}
\usepackage{amsmath,amssymb,amsfonts}
\usepackage[ruled, noend, linesnumbered]{algorithm2e}
\usepackage{algpseudocode}
\usepackage{graphicx}
\usepackage{textcomp}
\usepackage{xcolor}
\PassOptionsToPackage{demo}{graphicx}

\usepackage{bbold}
\usepackage{booktabs}
\usepackage{tabularx}
\usepackage{multirow}
\usepackage{amsfonts}
\usepackage{adjustbox}
\usepackage{lscape}
\usepackage[demo]{graphicx}
\usepackage{subfigure}
\usepackage{siunitx}

\def\BibTeX{{\rm B\kern-.05em{\sc i\kern-.025em b}\kern-.08em
    T\kern-.1667em\lower.7ex\hbox{E}\kern-.125emX}}

\usepackage{tikz}
\usetikzlibrary{shapes.geometric, arrows}

\tikzstyle{startstop} = [rectangle, rounded corners, 
minimum width=3cm, 
minimum height=1cm,
text centered, 
draw=black, 
fill=yellow!10]

\tikzstyle{io} = [trapezium, 
trapezium stretches=true, 
trapezium left angle=70, 
trapezium right angle=110, 
minimum width=3cm, 
minimum height=1cm, text centered, 
draw=black, fill=green!10]

\tikzstyle{process} = [rectangle, 
minimum width=3cm, 
minimum height=1cm, 
text centered, 
text width=3cm, 
draw=black, 
fill=orange!10]

\tikzstyle{decision} = [diamond, 
minimum width=3cm, 
minimum height=1cm, 
text centered, 
draw=black, 
fill=blue!10]
\tikzstyle{arrow} = [thick,->,>=stealth]
 
\begin{document}


\title{A 'MAP' to find high-performing soft robot designs: Traversing complex design spaces using MAP-elites and Topology Optimization}


\author{Yue Xie$^{1\dagger}$ and Josh Pinskier$^{2\dagger}$ and Lois Liow$^{2}$ and David Howard$^{2}$ and Fumiya Iida$^{1}$
\thanks{$^{1}$Yue Xie and Fumiya Iida are with Department of Engineering, University of Cambridge, Cambridge, CB2 1PZ, UK.}%
\thanks{$^{2}$Josh Pinskier, Lois Liow and David Howard with the Data 61, CSIRO, 1 Technology Ct, Pullenvale QLD 4069, Australia.}%
\thanks{$^{\dagger}$ Equal contributions as first authors}
}
\maketitle

\begin{abstract}
    Soft robotics has emerged as the standard solution for grasping deformable objects, and has proven invaluable for mobile robotic exploration in extreme environments. However, despite this growth, there are no widely adopted computational design tools that produce quality, manufacturable designs. To advance beyond the diminishing returns of heuristic bio-inspiration, the field needs efficient tools to explore the complex, non-linear design spaces present in soft robotics, and find novel high-performing designs.
    In this work, we investigate a hierarchical design optimization methodology which combines the strengths of topology optimization and quality diversity optimization to generate diverse and high-performance soft robots by evolving the design domain. The method embeds variably sized void regions within the design domain and evolves their size and position, to facilitating a richer exploration of the design space and find a diverse set of high-performing soft robots. We demonstrate its efficacy on both benchmark topology optimization problems and soft robotic design problems, and show the method enhances grasp performance when applied to soft grippers. Our method provides a new framework to design parts in complex design domains, both soft and rigid.
\end{abstract}


\section{INTRODUCTION}
The rapid rise of soft robotics, an emerging research field which uses soft and flexible materials to design robots which can deform around objects and reconfigure to match their environment, has created a paradox: evermore work is being produced by the field, (both academic and industrial), but few major design innovations are being generated. Most new designs are based on standard designs such as pneunets and tendon-driven continuum manipulators \cite{https://doi.org/10.1002/anie.201006464,Renda2014,Xavier2022}, which have now existed for over a decade. This situation can be attributed in part to the technological advancement within the field; however, the primary reason lies in the persistent difficulties that designers encounter when developing innovative soft robots \cite{Miriyev2020}. Soft robots are complex devices which exploit their material, structure and the environment to function effectively \cite{Pinskier_2023}. They operate \textit{with} rather than \textit{on} the environment, a critical difference which vastly increases the modelling complexity compared to traditional precisely configurable, rigid-linked robots. The complex interplay between materials, structure and behavior has proven difficult to resolve. As such no general purpose design tools exist to generate novel, physically-realisable, high-performing soft robots, with most new designs being created heuristically based on standard actuators \cite{pinskier2022bioinspiration}.

\begin{figure}[t]
    \centering
    \includegraphics[width=0.9\linewidth]{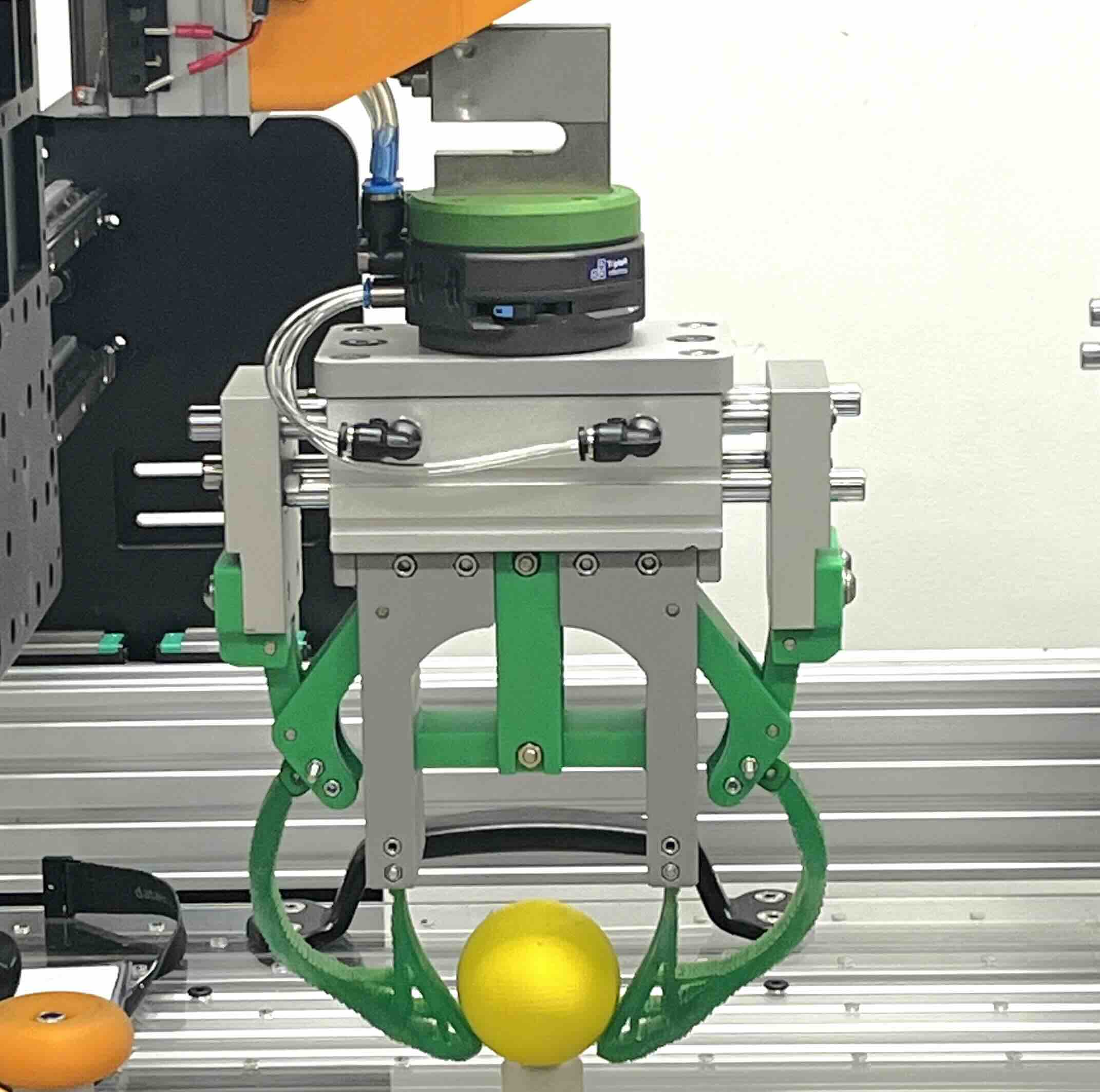}
    \caption{Example of high-performing soft robotic grippers generated through OIDD method.}
    \label{fig:soft_gripper}
\end{figure}

To address this problem, researchers have explored strategies including evolutionary computation~\cite{Cheney2014}, reinforcement learning~\cite{Wang2023, Schegg2023}, topology optimization (TO)~\cite{https://doi.org/10.1002/aisy.202300505,10122069,8481523,doi:10.1089/soro.2017.0121,10246335}, and other learning-based methods~\cite{Ma2021, Yao2023}. However, methods for learning complex designs (complete robot designs and actuation/gaits) in simulation are typically based on fast computer graphics engines. Their high-speeds necessitate abstracting away the real-world physics required to manufacture a functioning robot. In contrast, detailed optimization methods like topology optimization capture this physics in their accurate finite element method solvers, but use gradient-based solves which are unable to explore a complex design domain and find only a single design candidate. The selection of an appropriate design domain (bounding shape of design, design discretization/parameterization, initial guess) is an essential element in TO, particularly in highly non-linear and poorly understood problems. The challenge in these contexts is that the optimal choice for a design domain is often neither straightforward nor intuitive, and initial guesses must be close to the optimal solution for the solver to converge. We propose a detailed design optimization method which combines quality diversity methods with topology optimization (TO) to explore the design domain of TO problems, enabling diverse and high-performing design candidates to emerge, whilst retaining the detailed physics solvers, which enable designs both to be manufactured and to function in reality as they do in simulation (cross the reality gap). Our methodology enables systematic and generative exploration of the design domain, thereby mitigating human bias and enhancing design outcomes.

Topology Optimization~\cite{bendsoe1988,sigmund2013topology,eschenauer2001topology} is a mature design optimization technique, commonly used in structural engineering and architectural design. It is a powerful technology which has been used across multiple physical domains including thermal, structural, and fluidic problems \cite{prathyusha2022review,Picelli2020,Ikonen2018b} as well as physics-informed point-cloud reconstruction~\cite{Lowe2023}, but is most commonly used to design stiff-lightweight structures for automotive and aerospace applications~\cite{zhu2016topology}. Despite TO's remarkable achievements, it encounters challenges that stem from its focus on identifying single local optima, limiting exploration across varied design domains. This narrow focus restricts the generation of diverse design solutions critical for achieving optimal performance in complex and dynamic scenarios~\cite{mukherjee2021accelerating,jihong2021review}. These limitations highlight the need for innovative research methodologies that enable the design domain to evolve dynamically and intelligently throughout the optimization process, replacing the fixed design domain of traditional TO methods with a flexible one that adapts and reshapes during optimization. 

We present the Optimizing Initial Design Domain (OIDD) technique, a novel optimization methodology for soft robots which dynamically adjust the initial conditions and design domain of the TO solver to explore the space, using the placement of cavities to force the TO solver to find multiple, unconventional designs. It leverages MAP-Elites~\cite{DBLP:journals/corr/MouretC15}, a well-known quality-diversity algorithm, to navigate and exploit the high-dimensional design spaces. Through the dynamic adjustment of the initial design domain-by incorporating void regions of varying sizes and locations—OIDD facilitates the exploration of the design domain, fostering the emergence of diverse and high-performance structural configurations. An example soft gripper designed using the OIDD method is presented in Figure~\ref{fig:soft_gripper}, demonstrating the ability to design novel, functional soft-robots.The method is universal in nature and can be applied to topology optimization problems across physics and domains, however, here we focus on soft robotics, and demonstrate its efficacy in this challenging domain. 

Our main contributions are: 
\begin{enumerate}
    \item A novel evolutionary methodology to adapt the design domain during topology optimization
    \item A topology optimization formulation for multi-jointed inflatable grippers
    \item A set of diverse soft grippers, which outperform comparable benchmarks in grasp testing
\end{enumerate}

\section{Enhancing Structural Design through Optimized Initial Domains}
\label{sec:method}
\begin{algorithm}[t]
    \caption{OIDD method}
    \label{algorithm}
    \For{iteration $\in [1, I]$}{
    \If{First iteration}{
    $\mathcal{B}\leftarrow$ random initial solutions
    }
    \Else{
    $\mathcal{B}\leftarrow$ select solutions from archive $\mathcal{A}$
    }
    $\tilde{\mathcal{B}} = (\tilde{x_j})_{j \in [1,N]} \leftarrow $ new solutions(undergo variations) \;
    \For{$j \in [1, N]$}{
    run \textit{SIMP}, get return $F(x_j)$ and $M(x_j)$ \;
    cell $\leftarrow$ get grid cell of descriptor $\tilde{d}(M(x_j))$\;
    $x_{cell} \leftarrow$ get content of cell\;
    \If{$x_{cell}$ is None}{
    Add $x_{j}$ to cell
    }
    \ElseIf{$F(x_j)< F (x_{cell})$}{
    Replace $x_{cell}$ with $x_j$ in cell
    }
    \Else{
    Discard $x_j$
    }
    }
    }
    \Return archive $\mathcal{A}$
\end{algorithm}

In this section, we present the development of the Optimized Initial Design Domain (OIDD) method and its integration with the Solid Isotropic Material with Penalization (SIMP) framework.
At its core, the SIMP method discretizes a design domain into a set of elements, representing the material. The elements are allowed to occupy continuous distribution from $0$ to $1$, with $0$ representing a void region and $1$ being solid material. For convenience the design domain is typically a rectangle in 2D (cuboid in 3D) and divided into a set of square (quadrilateral) or cube (hexahedron) elements. Although convenient, this is likely a sub-optimal configuration, as the design domain strongly influences the resulting designs in TO problems. A detailed review of the SIMP formulation is beyond the scope of this work, but its formulation is well-established and can be found in ~\cite{bendsoe1989optimal, Bendsoe2003}.

Addressing the limitations of conventional TO, the OIDD strategy employs the MAP-Elites algorithm to introduce variability in the initial design stages. Incorporating voids of differing sizes and placements within the design domain enables a comprehensive exploration of potential configurations and their impacts on structural performance. Whilst a pure evolution driven optimization is appealing, it leads to non-physical solutions with disjoint regions~\cite{Guirguis2020}. Our methodology ensures a balanced examination of design alternatives, optimizing for structural integrity and material efficiency.

\subsection{OIDD Algorithm}
The algorithm framework, for the method is outlined in Algorithm~\ref{algorithm}, its detailed function is as follows:
\begin{enumerate}
    \item \textbf{Initialization of Design Domains} We begin by generating an array of initial design domains, each featuring unique void region configurations. This step lays the groundwork for extensive design exploration by ensuring a diverse set of starting points.
    \item \textbf{Archive Creation Using MAP-Elites} The MAP-Elites algorithm creates an archive ($\mathcal{A}$), a multi-dimensional grid, where each cell in the grid represents a unique combination of design features (descriptors) such as the location and size of void regions. This categorization promotes diversity by ensuring different design configurations are explored and stored.
    \item \textbf{Topology Optimization and Performance Evaluation}  Each initial design domain undergoes topology optimization using the SIMP methodology. The performance of each optimized design is then evaluated based on problem-specific objectives, such as structural compliance.
    \item \textbf{Integration into MAP-Elites Grid}  Each optimized design is placed into the appropriate grid cell in the MAP-Elites archive ($\mathcal{A}$) based on its descriptors. If a grid cell already contains a design (an elite), the new design replaces the current elite if it has better performance. This ensures that only the best-performing designs for each descriptor combination are retained.
    \item \textbf{Domain Evolution} The design domains are evolved with each new generation by applying evolutionary operators (such as mutation and crossover) to the void region configurations. This gradual improvement in the positioning and sizing of voids encourages continuous design enhancement.
\end{enumerate}

Steps (3)-(5) are then repeated until the computational budget (maximum number of iterations) is reached. Throughout this iterative process, the MAP-Elites archive ($\mathcal{A}$) is continuously updated, storing and categorizing the best designs based on their descriptors. The resulting archive showcases a diverse set of high-performing designs, each representing a unique combination of features.

\subsection{Encoding strategy of initialization of design domain}

\begin{figure}[t]
    \centering
    \subfigure[]{\includegraphics[width=0.45\linewidth]{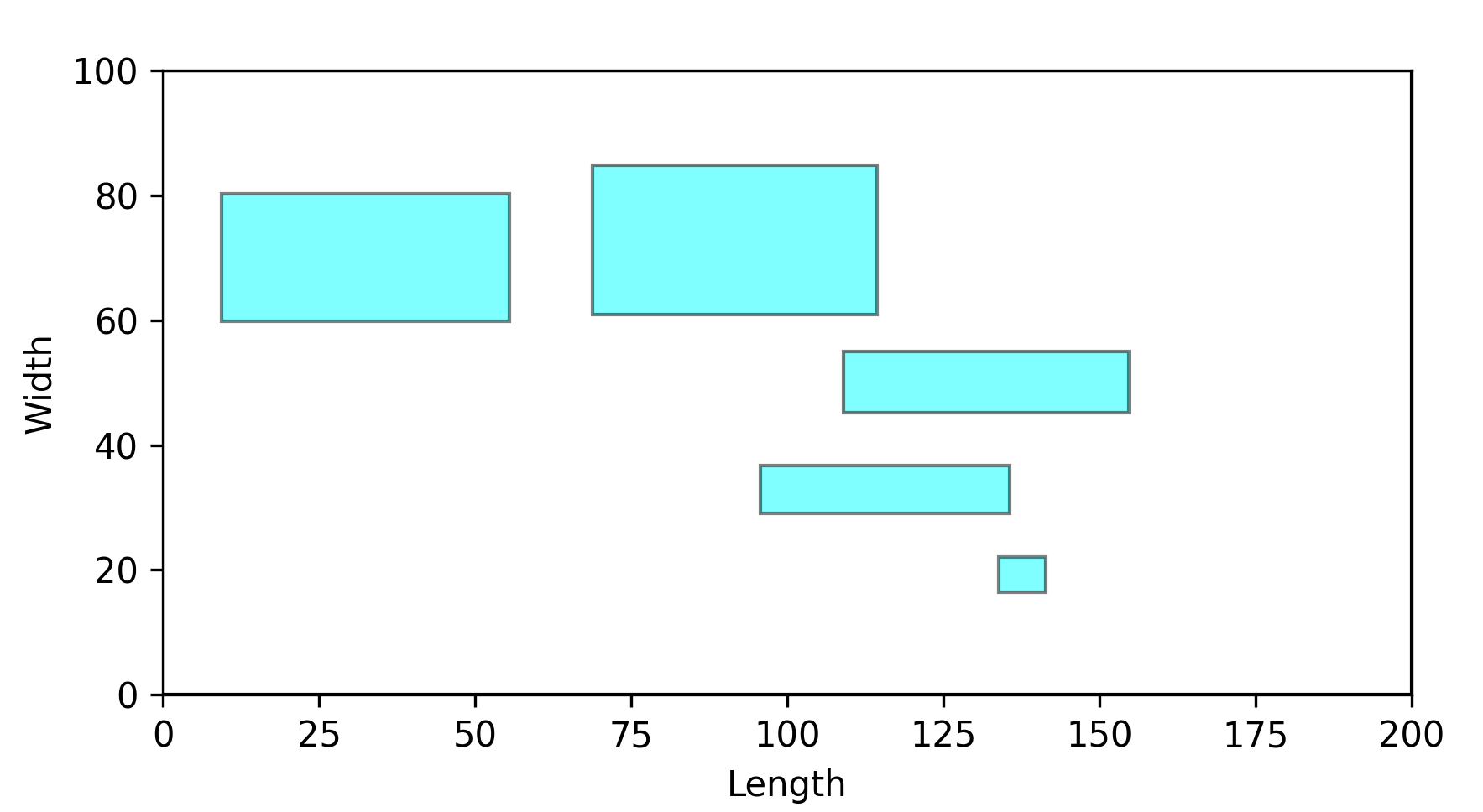}
    \label{fig:designdomain_2D}}%
    \subfigure[]{\includegraphics[width=0.45\linewidth]{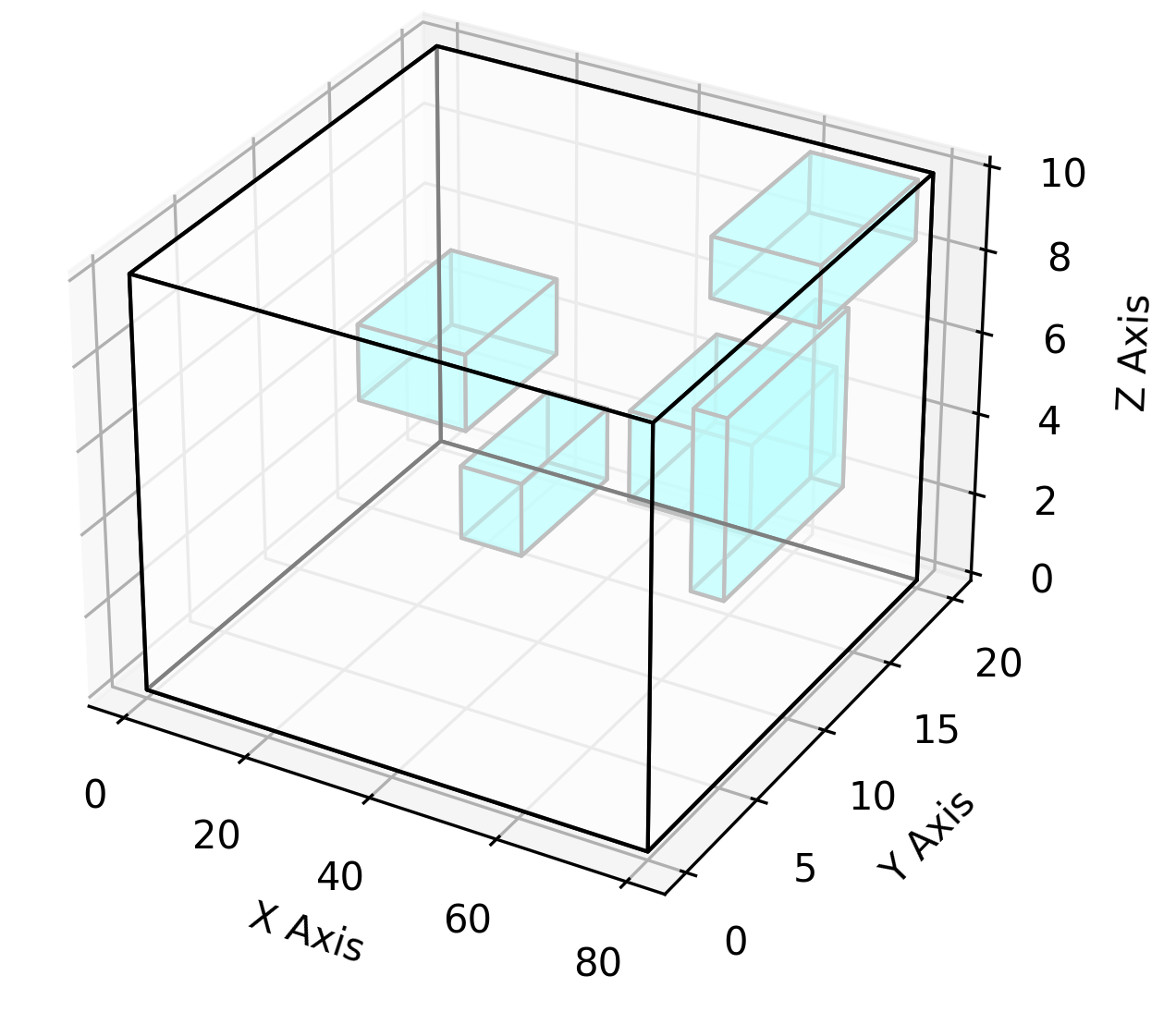}
    \label{fig:designdomain_3D}}%
    \caption{Illustration of the encoded design domain initialization. a) 2D Design Domain Initialization b) 3D Design Domain Initialization}%
    \label{fig:designdomain}%
\end{figure}

The initialization of the design domain impacts the strategic implementation of the OIDD method. Our approach refines the traditional TO design process by integrating decision variables that define void configurations, including the number of void regions, locations, and dimensions within the initial design domain. Introducing the number of void regions as a variable adds depth to our analysis, allowing for a detailed investigation into how variations in void quantity influence structural durability and functionality. Specifying each void's position and size further enables examining their collective impact on load distribution and stress endurance across the structure. Figure~\ref{fig:designdomain} illustrates the design domain initialization within OIDD methodology. In Figure~\ref{fig:designdomain_2D}, we see a 2D representation of the design domain, where void regions are distributed in varying sizes and locations, demonstrating the initial setup's adaptability and diversity.  Figure~\ref{fig:designdomain_3D} extends this concept into three dimensions, highlighting how void activation—depicted by the presence or absence of cyan blocks within the structure—plays a crucial role in defining the design's structural integrity and potential functionality. 

We apply an encoding strategy that represents the decision variables as a concatenated string that enhances the granularity and flexibility of design representation. This encoding strategy employs a five-bit string denoted as $v= \{x,y,l,w,a\}$ for each potential void region, designed to encode the top-left vertex coordinates $(x,y)$, the dimensions of the void (length $l$ and width $w$), and a binary indicator $a$ denoting the void's activation within the design domain. The fifth bit of each void region is crucial for whether the void region actively contributes to the design ($1$) or remains latent ($0$), thus enabling a dynamic exploration of the design space by selectively activating or deactivating void regions. 

This encoding strategy establishes a robust framework for systematically investigating diverse design configurations. In each optimization, we set the maximum available number of void regions, $n$. Therefore, the length of the decision variables is $5n$ for 2D TO problems and $7n$ for 3D problems. 

\subsection{Diversity Feature selection}
To generate a diverse pool of designs, a set of features are required which capture the designs' unique properties and differences. Here we use two features to quantify diversity arising from void configuration: Shannon entropy and void-centroid distance.

Shannon entropy indicates the distribution of void sizes within a design, providing a quantitative assessment of uniformity in material distribution. To evaluate this feature for a set of decision variables, we first calculate the volume of each void region based on its dimensions (length, width, and height, as specified by the decision variables) and normalize these volumes relative to the total volume of all voids within the design. Then, we apply the Shannon entropy formula $H = -\sum (p_i \log p_i)$, where $p_i$ is the normalized volume of the $i$-th void. This calculation transforms the distribution of void volumes into a single metric that encapsulates the diversity of void sizes, offering insights into the design’s balance between structural strength and material usage.

Beyond just void size, the spatial arrangement of voids significantly influences structural performance. To evaluate the diversity of this feature, we use a distribution metric that examines the distances between void centroids, assessing the strategic dispersion of voids. This metric highlights the clustering or uniform spread of voids and aids in understanding their collective impact on structural performance.
We calculate the centroid based on each void's top-left vertex and dimensions. Then, we compute the Euclidean distances between each pair of void centroids to measure the spatial dispersion of voids within the design domain. Finally, we employ statistical measures (mean, variance, and standard deviation) to quantify the spatial distribution. A lower variance indicates a more uniform distribution of voids, while a higher variance suggests clustering or irregular dispersion.

Together, these two features provide a multidimensional characterization of design possibilities within the MAP-Elites archive. Our strategy captures a broad spectrum of design diversity, ensuring that each solution within the MAP-Elites archive is distinct and informative, expanding the horizons of what is achievable in topology optimization.

\section{Evaluation on Benchmark Problem}
\label{sec:exp}

\begin{figure}[t]
    \centering
    \includegraphics[width=0.85\linewidth]{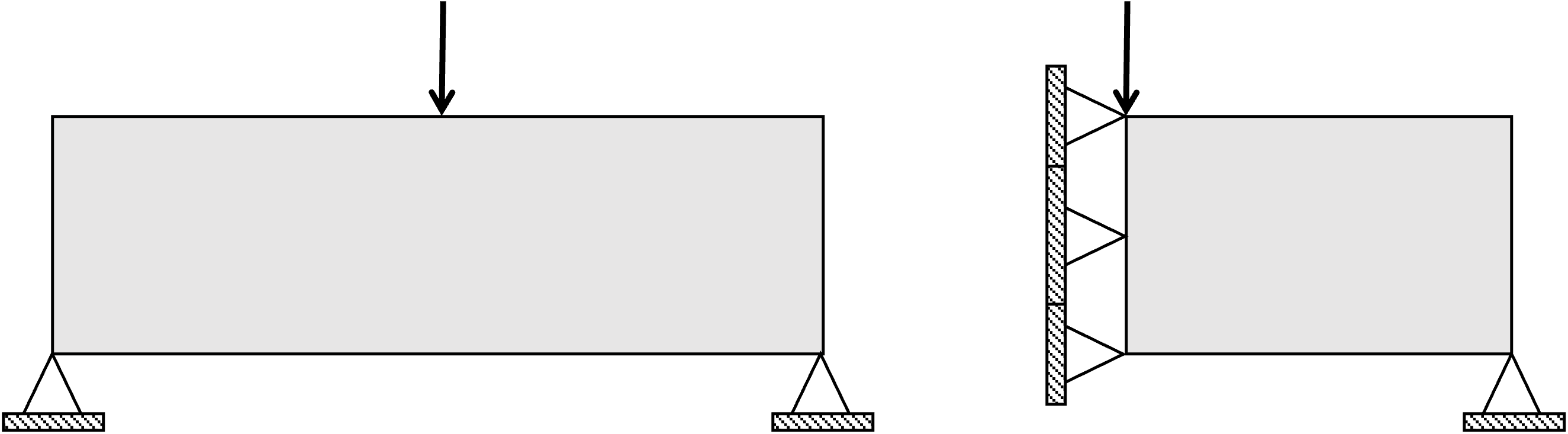}
    \caption{Loading and boundary conditions of \textbf{MBB beam}. Left: full design domain, right: half design domain with symmetry boundary conditions.}
    \label{fig:mbb}
\end{figure}

\subsection{The MBB Problem}
In this section we evaluate the performance of the Optimized Iterative Design Diversification (OIDD) methodology on a benchmark TO problem, the MBB beam \cite{DBLP:journals/corr/abs-2201-11513}.
While the problem has been thoroughly explored within the TO community, many high-performing configurations exist. Hence, beyond demonstrating the validity of the OIDD method, finding diverse MBB designs is also of interest to researchers and practitioners. 

The MBB problem is a structural compliance problem, similar to designing a bridge. The optimization problem aims to minimize the structural compliance of the beam subject to a material volume constraint, hence maximising stiffness for a given mass. The initial design domain and boundary conditions for this problem are depicted in Figure~\ref{fig:mbb}.

This example operates under a fixed termination condition set at $50$ OIDD (outer-loop) iterations. The initial population for each run is set at $20$, with the algorithm producing $10$ offspring in every iteration. This configuration ensures a balanced approach between exploration and exploitation within the design space. Each design is topology optimized for $50$ iterations in an inner-loop, and the design domain contains $200 \times 100$ elements, with upto $10$ void regions.

\begin{figure}[t]
\centering
\subfigure[]{\includegraphics[width=0.8\linewidth]{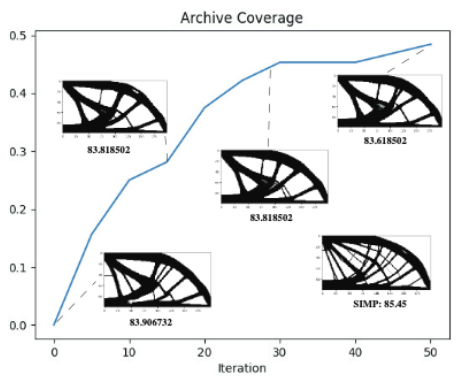}
\label{fig:mbb_iteration}}
\subfigure[]{
   \includegraphics[width=0.8\linewidth]{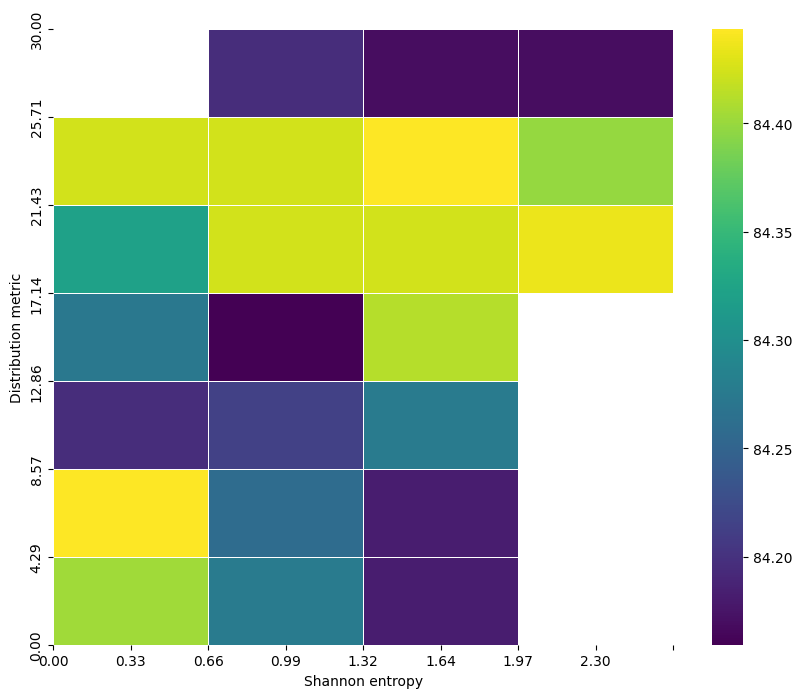}
  \label{fig:mbbheat}}
\caption{Analysis of MBB Beam Optimization. a) Evolution of archive coverage and corresponding optimal designs. b) Distribution of high-performing solutions in the behavior space of MBB beam}
\label{fig:MBBresults}
\end{figure}

\subsection{MBB Results}
The results of the OIDD optimization on the benchmark MBB problem are presented in Figure~\ref{fig:MBBresults}, showing the growth of the archive and the overall best design as the optimization progresses (Figure~\ref{fig:mbb_iteration}), and the final archive along with selected design (Figure~\ref{fig:mbbheat}). Overall 20 elite designs were found in the 64 cell archive.

Initially, the archive expands quickly as every cell is empty and hence any valid entry qualifies as elite. However there are diminishing returns as the cells fill up, making it harder and harder to find valid and elite entries This could indicate that the solution space has been thoroughly explored or that the algorithm's parameters are tuned towards intensification over diversification as the search progresses.
The MBB design resulting from a standard topology optimization run is shown in the bottom-right in Figure~\ref{fig:mbb_iteration} (random domain initialization without any void regions). The other four designs are the best performing designs chosen from the archive corresponding to their iterations. The numbers accompanying each design represent the compliance objective (lower is better). Evidently, several distinct designs can be found which outperform standard TO.

Figure~\ref{fig:mbbheat} complements this by showing the distribution of high-performing solutions in the behavioral space of the MBB beam. Each cell in the heatmap represents a unique combination of diversity measures values, with the color indicating the objective value of the optimal solutions. The skewed distribution in the heatmap, with a concentration of viable designs within certain regions, is likely a function of the underlying physics and constraints imposed on the beam's design, where only specific configurations can meet the performance criteria. 

For statistical validation of the OIDD method's efficacy, the Kruskal-Wallis test with a $95\%$ confidence interval and the Bonferroni correction method are employed. The comparison between standard SIMP and OIDD-enhanced SIMP shows a statistically significant difference in performance. The average objective value for standard SIMP over $30$ independent runs is $84.62$ with a standard deviation of $0.81$, while for OIDD+SIMP, the average objective is $82.30$ with a standard deviation of $0.33$. This statistically significant difference highlights the enhanced efficiency and effectiveness of the OIDD approach in topology optimization.

\section{Application to Soft Robotics}
\label{sec:SR-exp}
The previous MBB beam example demonstrated the viability of the OIDD method to a well-known, benchmark TO problem. Here, we demonstrate its application to soft robotics.
In the MBB and other classical TO problems, the design goal can be readily transformed into a measurable cost function and constraints (e.g minimizing stiffness subject to weight costraint). Soft robots, which need to be flexible but strong and light but powerful, are challenging to formulate into a tractable and meaningful, scalar cost function.
In this section we demonstrate the value of the OIDD method in two typical soft robotics applications, soft grippers and pneumatic soft actuators. We demonstrate the ability of our method to overcome local minima and generate diverse sets of high-performing candidates for further detailed evaluation and integration.

\subsection{Bending Soft Gripper}

\begin{figure}[th]
    \centering
    \includegraphics[width=0.75\linewidth]{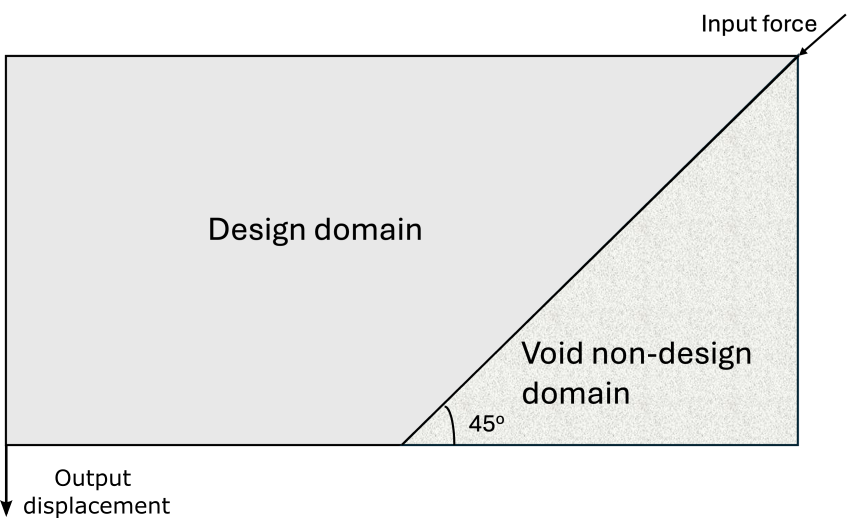}
    \caption{Design domain of soft gripper optimization: the finger should bend inwards from the bottom left corner when a force is applied at the top right. The void space is intended to maintain a clear path for the gripper to close in a 2 fingered configuration}
    \label{fig:gripper}
\end{figure}

\begin{figure}[th]
    \centering
    \subfigure[]{
    \includegraphics[width=0.8\linewidth]{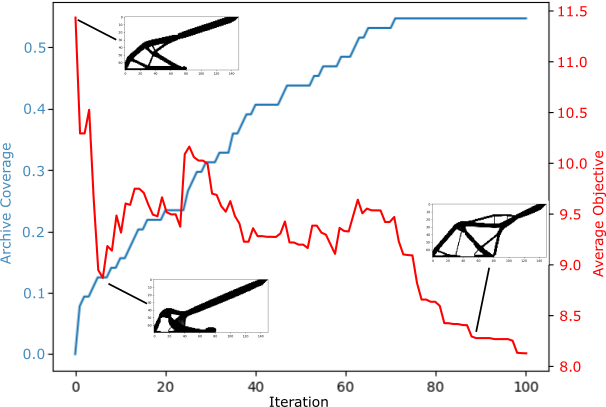}
    \label{fig:gripper_iteration}}
    \subfigure[]{
    \includegraphics[width=0.8\linewidth]{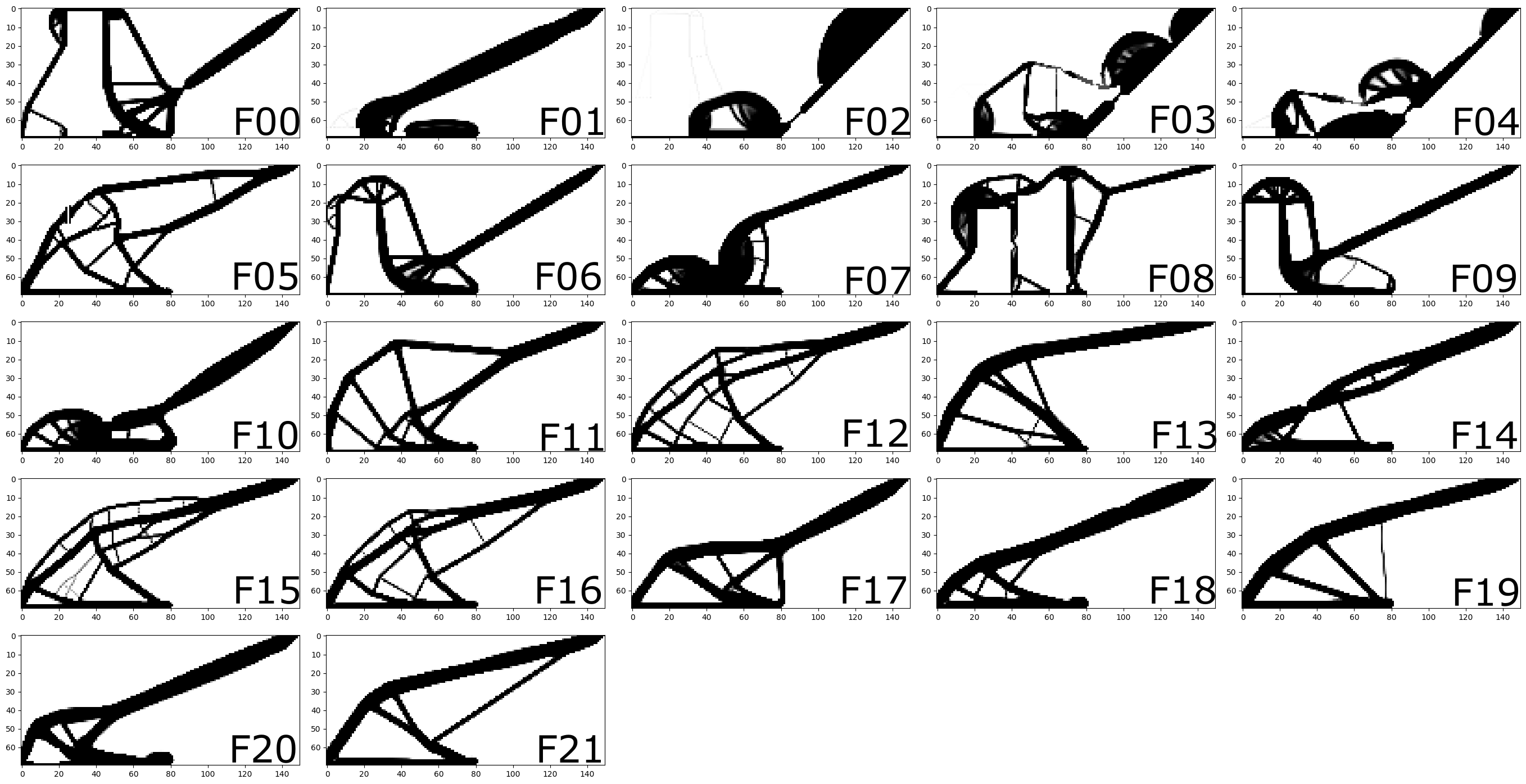}
    \label{fig:digital}}
    \caption{Optimization results of soft gripper problem: a) optimization performance, showing evolution of the archive coverage and average objective value of the designs produced in each generation, and the corresponding best designs at selected points  b) Optimal designs of the soft gripping bending finger obtained by OIDD}
    \label{fig:Gripperresults}
\end{figure}

We first optimize the design of a soft
gripping finger (Figure~\ref{fig:soft_gripper}), which bends around an object when one side of the gripper is translated relative to the other. The bending soft gripper problem is based on~\cite{doi:10.1089/soro.2017.0121}. It applies a force to one corner of the design domain, creating a compressive load on the soft gripper, that forces it to bend inwards. This design is of interest because it is an implementation of soft grasping with published optimized designs, and because it has been shown that the resulting designs can be improved by manually tuning the design domain \cite{10246335}. The design domain of the problem is presented in Figure~\ref{fig:gripper}. It is parameterised by $150 \times 70$ elements. Here the OIDD algorithm is run for 100 iterations, with a population size of 10.

\subsubsection{Numerical Results}

The experimental results, presented in Figure~\ref{fig:Gripperresults}, illustrate the efficacy of OIDD in generating a suite of optimized topologies tailored to the unique demands of compliance in robotic grippers. Figure~\ref{fig:gripper_iteration} traces the evolution of archive coverage and average objective value (average of population in each iteration) over iterations. The plot shows a steady increase in archive coverage as the method progresses, highlighting the method's thorough exploration of the design domain. However, it plateaus at 55\%, as resulting regions are increasingly challenging to find or infeasible.
Whilst not a monotonic improvement, on average the optimizeer finds increasingly high quality solutions as it progresses, improving from an average objective of $11.5$ in the first iteration to $8.2$ in the last. 
The best compliance metric seems to gradually plateau, indicating the convergence towards an optimal solution in the middle of the process.
 
Figure~\ref{fig:digital} presents the resulting optimized topologies of the compliant gripper mechanism. Notably, the designs reveal a variety of structural adaptations driven by the constraints imposed by void regions, indicative of the algorithm's responsiveness to different initial conditions and constraints.

\subsubsection{Experimental Grasp Testing}
\begin{figure}%
    \centering
    \subfigure[]{\includegraphics[width=1.0\linewidth]{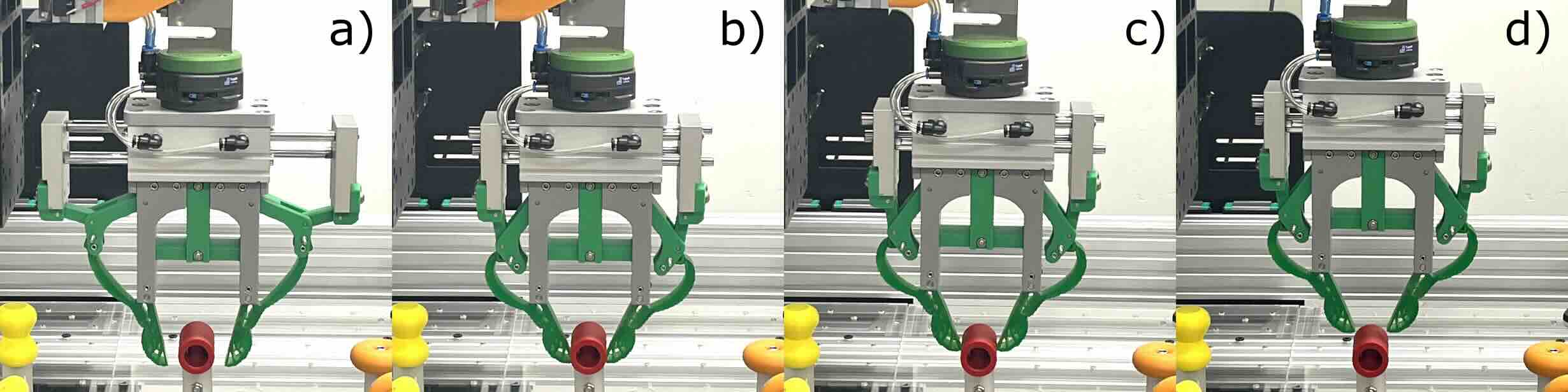}
    \label{fig:grasp_cycle}}%
    \vspace{2pt}
    \subfigure[]{\includegraphics[width=1.0\linewidth]{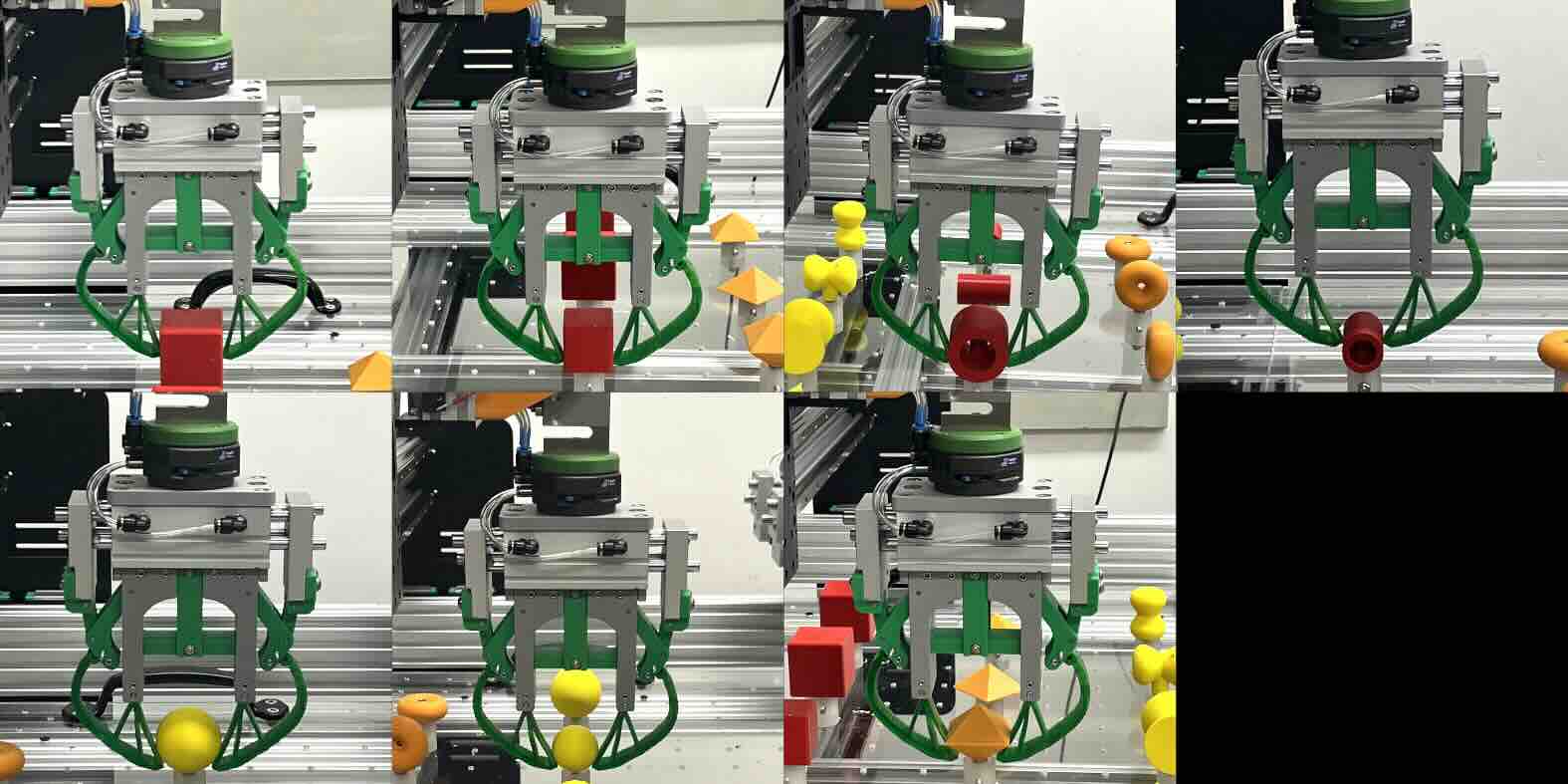}
    \label{fig:object_set}}
    \caption{Experimental grasping procedure and platform. a) Testing cycle: for each object/gripper pair the gripper is lowered around the object, closed, then slowly raised until contact is lost. The forces are recorded throughout. b) Set of grasped objects shown during grasping by F10 gripper. Top row (Left to right): \SI{40}{\milli\meter} cube, \SI{25}{\milli\meter} cube, \SI{40}{\milli\meter} diameter tube, \SI{25}{\milli\meter} diameter tube. Bottom row: \SI{40}{\milli\meter} diameter sphere, \SI{25}{\milli\meter} diameter sphere, \SI{40}{\milli\meter} diamond }%
    \label{fig:setup}%
\end{figure}

The topology optimization formulation creates high-performing bending \textit{soft fingers}, however what we are really interested in is \textit{soft grippers}. Because of the sharp discontinuity that occurs when grippers make contact with an object, optimizing soft grippers in a realistic grasp environment is an intractable problem. Hence, the problem is abstracted into a function of shape in free-space (i.e maximising bending), as an approximation of the desired grasp behavior.
To evaluate the designs' true grasp performance, we measure the peak grasp strength (retention force) of a representative subset (9 fingers) of the resulting designs. The designs are evaluated using an automated gantry platform, which contains a set of stationary, rigid 3D printed test objects. The platform uses a pneumatic cylinder to open and close candidate soft fingers around the objects. A custom 2-bar mechanism is used in this work to convert the horizontal motion to the desired \SI{45}{\degree} path. The mechanism and grasp process are shown in Figure~\ref{fig:grasp_cycle}. In each test, the fingers are lowered around the fixed object, closed to induce contact and then raised until contact is broken, with grasp (vertical) force recorded throughout. 7 objects are used in the experiment (Figure~\label{fig:object_set}) giving a range of shapes and sizes including smooth spheres and jagged diamonds. The fingers are printed using Agilus 85 (Shore 85A) resin on a Stratasys J850 polyjet printer, and have dimensions $75 \times 37.5 \times 10 \si{\milli\meter}$

\begin{figure}[th]
    \centering
    \subfigure[]{\includegraphics[width=1.0\linewidth]{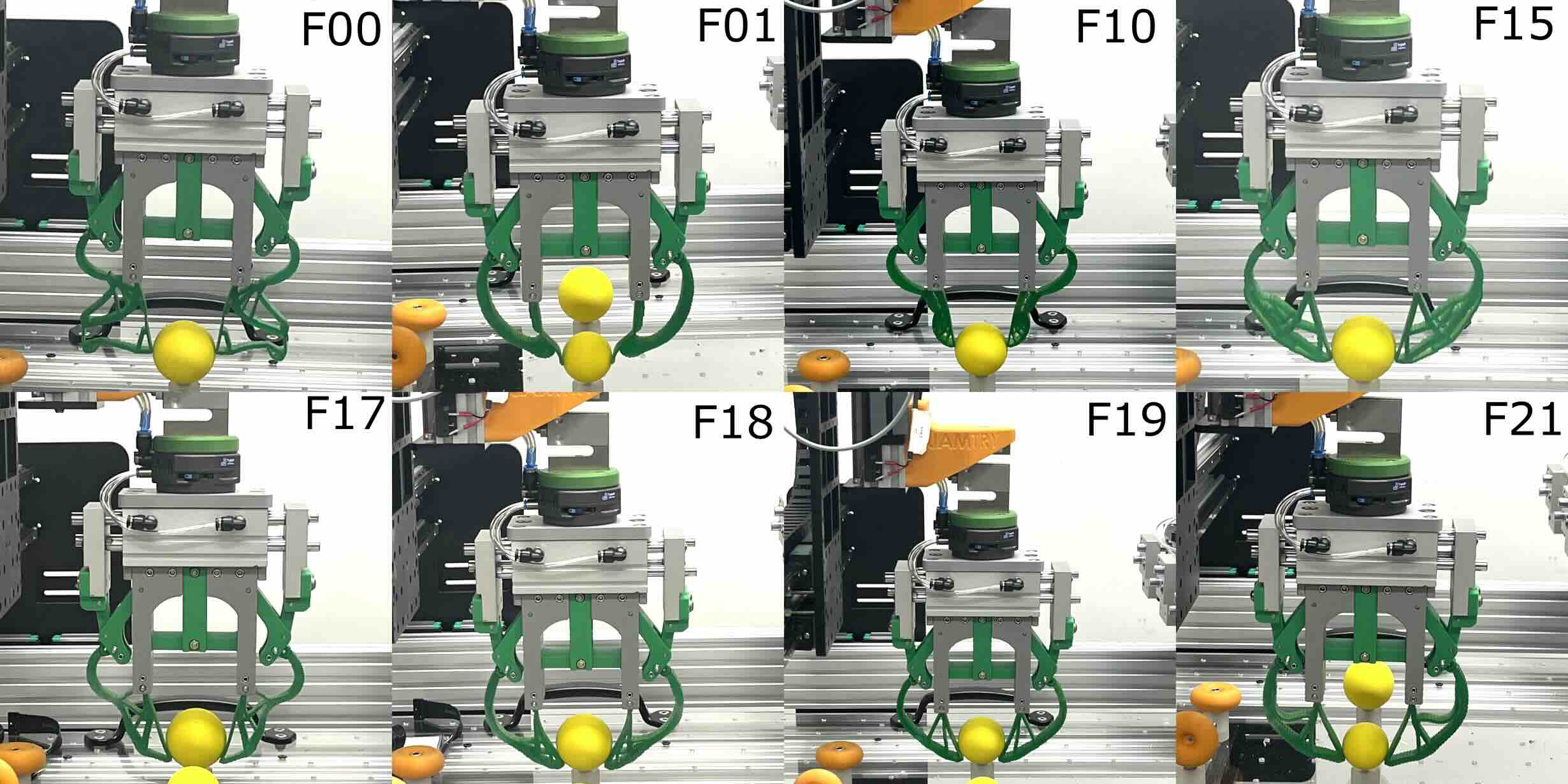}
    \label{fig:grasp_fig}}
    \subfigure[]{\includegraphics[width=1.0\linewidth]{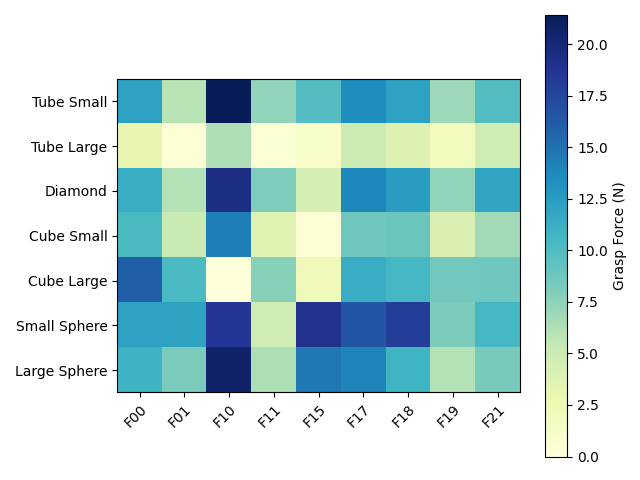}
    \label{fig:grasp_heatmap}}
    \caption{Experimental Grasping results. a) Sample of grippers for testing, illustrating grasping modalities b) Complete set of grasped objects c) Heatmap of maximized Grasp Force Measurements across Various Objects and Soft Gripper Designs}
    \label{fig:printheat}
\end{figure}

The grasping mechanism of the grippers and the resulting heatmap of grasp strength is shown in 
Figure~\ref{fig:printheat}. Several unique grasping mechanisms emerge from the OIDD optimization: Designs F11, F15, F18, F19 and F21 all have long linkages connecting the gripping region to the input force (2-bar mechanism), these curve significantly when actuated, allowing the gripper to bend around the object with a large contact area but relatively low force (due to strain energy dissipating in the structure). F01 has a similar mechanism, but uses a soft pad for gripping rather than a reinforced truss 'jaw'. Finally, F00, F10 and F17 have a shorter, stiffer linkage, which gives a larger force at the tip for strong pinch-grasping.
It can be seem that the pinch-grasps typically give stonger grasps than those that envelop the object. However the result is not universal, with also F18 giving strong grasps (and large displacement). Both F10 and F18 are excellent designs for future use, with the choice depending on the application and its requirements.
Interestingly almost every gripper tested outperforms F19, a design which is nearly identical to the published benchmark \cite{doi:10.1089/soro.2017.0121}, highlighting the efficacy of our method.

\subsection{3D Pneumatic Soft Finger}
 As a final example, we demonstrate the application of OIDD to a 3D pneumatic soft finger, similar to widely used Pneunets \cite{https://doi.org/10.1002/anie.201006464}. Inflatable fingers have been topology optimized in the past \cite{pinskier2022bioinspiration,Chen2020,10122069}; given the strongly nonlinear behavior of these fingers, and their sensitivity to small design-changes, it is likely that these are sub-optimal solutions, which can be improved using OIDD.

\begin{figure}[th]
    \centering
    \subfigure[]{\includegraphics[width=0.9\linewidth]{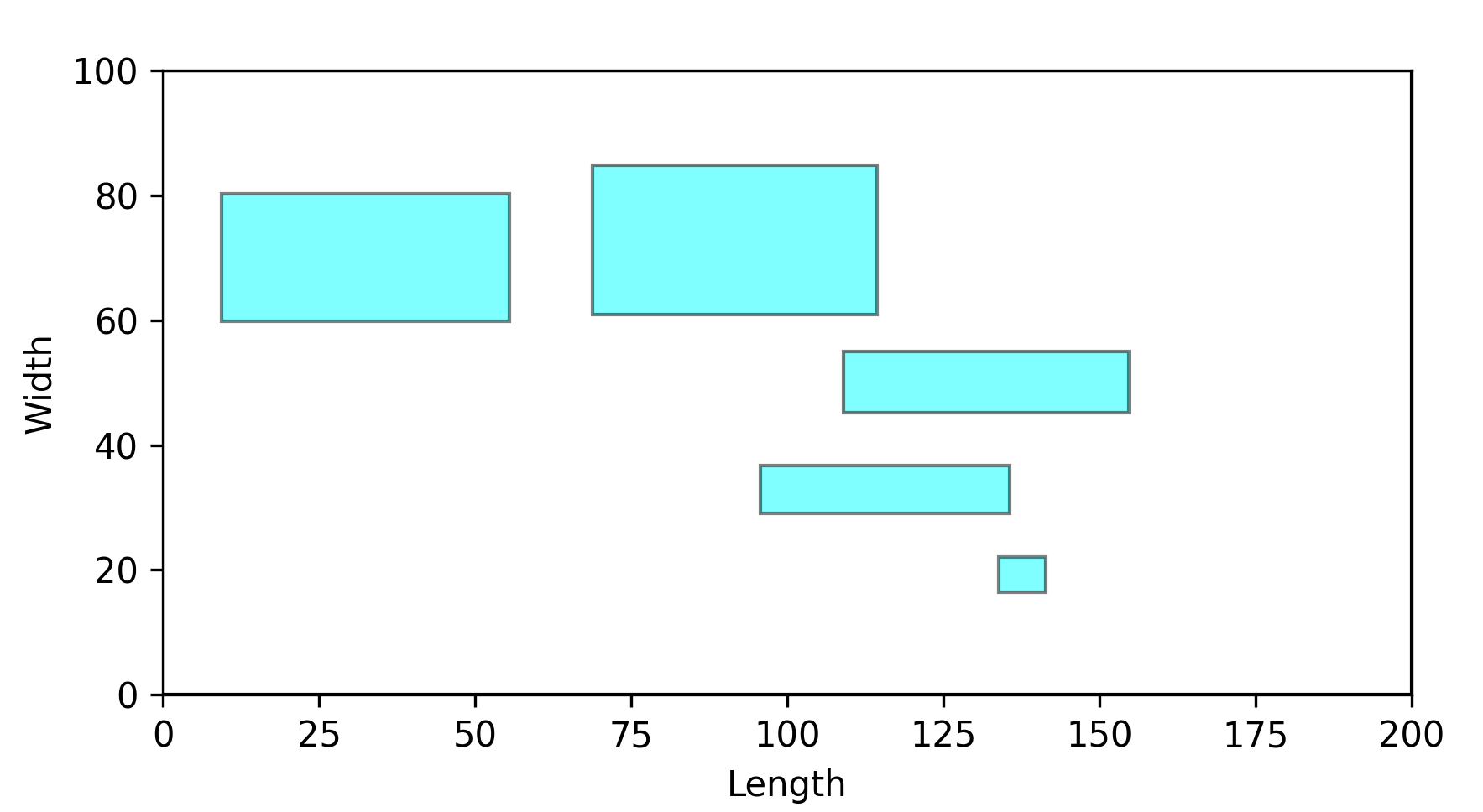}}
    \caption{optimized pneumatic soft finger example. a) Design Domain of soft finger joint b) Performance of representative group of results in high-fidelity simulation.
Left side: optimized single links generated by OIDD. Right: Simulated inflation of corresponding soft fingers in 3 link configuration of \SI{0.5}{\mega\pascal} rubber with \SI{0.5}{\mega\pascal} inflation pressure}
    \label{fig:Softfinger}
\end{figure}

To investigate this problem, we investigate the optimization of multi-jointed soft fingers, using an approach similar to \cite{8481523},
The design domain of the multi-jointed soft finger is shown in Figure~\ref{fig:Softfinger}(a), and is paramaterised by $20 \times 20 \times 10$ \SI{1}{\milli\meter} elements. The finger is intended to bend when pressure is supplied to a central tube. As the tube expands, the surrounding material should steer the output downwards.

The OIDD process was run for 10 generations with a population size of 10, and resulted in 24 valid designs on the QD grid. 
A sample of the resulting designs is presented in Figure~\ref{fig:Softfinger}(b), along with their displacement when simulated as a 3 jointed finger in Comsol Multiphysics. Unlike the original optimization, the Comsol simulation captures both the geometric and loading non-linearities present in soft fingers.
Two interesting features are evident in the results: firstly, the OIDD method has learned to 'checkerboard' the designs. This phenomenon, in which the optimizer converges on an alternating pattern of filled and empty voxels, is common within TO. It exploits a flaw in the FEM simulation to reduce material usage whilst retaining stiffness.
Although not typically a desirable outcome, finding such a solution demonstrates the utility of OIDD in illuminating 'high-performing' solutions even within large search spaces.
Secondly, we see the sensitivity of the performance to small changes in design. When using a high fidelity solver, small changes in the design not only vastly change the output, but can even invert the displacement.
Such complexity highlights the need for our multistage optimization approach that overcomes nonlinearities and local minima to find high-quality candidates for further evaluation. Capturing the full physics of soft robotics in a fast simulator is an ongoing challenge for the field, however even using coarse simulation and a highly abstracted problem, we generate a number of high-quality results.

\section{Conclusion}
\label{sec:con}

In this research, we introduced the Optimized Initial Design Domain (OIDD) methodology, which significantly advances topology optimization. This novel hybrid design optimization method exploits global-search evolution and fine-grained topology optimization to find diverse and structurally efficient designs. By strategically adjusting the initial design domain, our methodology facilitates the generation of solutions that are both structurally efficient and considerate of material constraints.

In benchmark optimization problems and sim-to-sim and experimental evaluation in soft robotic design problems, we show that OIDD finds higher-performing designs than SIMP alone and that the resulting designs outperform benchmarks in physical experimental evaluation. Physical testing is essential to our research, as the substantial correlation between simulated optimization results and experimental outcomes highlights the OIDD method's robustness and capacity to bridge the gap between simulation and reality reliably.

In future work, the OIDD methodology may be further refined to streamline the design process for soft robots, leading to the innovation of soft robotic design. The impact of the OIDD method on this field is anticipated to open up unprecedented opportunities in how soft robots are imagined and utilized.

\section*{Acknowledgement}
This project has received funding from the European Union’s Horizon 2020 research and innovation programme under the Marie Skłodowska-Curie grant agreement No 101034337.

\bibliographystyle{IEEEtran}
\bibliography{IEEEabrv,bib}








\end{document}